# Modular Cognitive Architecture Emerges in Large Language Models

**Pengrui Han, Jacob Andreas, Evelina Fedorenko[†], Andrea Gregor de Varda[†]**

Massachusetts Institute of Technology

{phan3, jda, evelina9, devar_ag}@mit.edu

[†]Equal contribution

Code & Data Project Website

**Abstract**

The human brain exhibits a striking degree of functional specialization, with distinct networks supporting language, formal reasoning, reasoning about other minds, and reasoning about the physical world. Is this modular organization a fundamental principle of how intelligent systems must be built, or an evolutionary accident specific to biological brains? Here, we test whether a similar organization emerges in Large Language Models—another class of intelligent systems created through a very different optimization process. Using circuit analyses across N=46 tasks spanning four cognitive domains (language, formal reasoning, social reasoning, physical reasoning), we find that LLMs develop a modular architecture that mirrors the human brain: tasks drawing on the same network in humans recruit overlapping neurons in LLMs, whereas tasks drawing on different networks recruit distinct neurons. The convergent emergence of modularity in brains and neural networks suggests that it may be a fundamental property of intelligent systems.

## Introduction

For centuries, neuroscientists have studied a single instance of a complex intelligent system—the brain—uncovering its numerous structural and functional properties. For example, brains of all vertebrates consist of two hemispheres. The two hemispheres are largely but not fully symmetrical anatomically and functionally (Toga & Thompson, 2003;

Kong et al., 2018). Some functions are lateralized: language, for instance, is processed primarily in the left hemisphere in most people (Fedorenko et al., 2024a). Moreover, at least primate brains show a substantial degree of functional specialization, with different cortical regions selectively engaged in distinct mental processes, from the visual recognition of faces and places (Kanwisher et al., 1997; Epstein et al., 2001), to auditory processing of speech sounds and music (Norman-Haignere et al., 2015), to more abstract domains of high-level cognition (Saxe & Kanwisher, 2003; Fedorenko et al., 2011). Which of these properties are critical for human intelligence and which are mere accidents of biological evolution? For some properties, variation across humans rules out a critical role in intelligence: for instance, although language is typically left-lateralized, in some individuals it is right-lateralized with no apparent cost to language ability (e.g., Knecht et al., 2001). Similarly, having two hemispheres does not appear to be strictly necessary for intelligent behavior: individuals who undergo hemispherectomy (removal of a hemisphere) early in life to treat epilepsy can develop close-to-normal cognitive function with a single hemisphere (e.g., Kliemann et al., 2019). For functional specialization, however, humans show little variation: virtually every neurotypical brain exhibits the same broad modular organization (Kanwisher, 2010; Fedorenko et al., 2024a). This consistency across humans makes modularity a particularly compelling candidate for a feature that may play a central role in intelligence.

A strong test of whether a feature is broadly favored for a particular behavior is whether it independently emerges in multiple systems facing similar computational demands. In biology, evolution of a trait in unrelated or distantly related species (e.g., echolocation in bats and cetaceans, or wings in bats and birds) is taken as evidence that this feature is adaptive, driven by the nature of the problem and the structure of the environment. The recent emergence of *in silico* intelligent systems, Large Language Models (LLMs), offers a new opportunity to ask whether certain properties of biological brains are solutions that any form of optimization towards intelligent systems will tend to discover. LLMs produce human-like language (Jones & Bergen, 2025) and capture human behavioral and neural responses to language (Schrimpf et al., 2021; Caucheteux & King, 2022; see Tuckute et al., 2024 for review). Furthermore, the most advanced LLMs achieve impressive performance on cognitive tasks (Guo et al., 2025) and explain human behavior in domains such as reasoning and social inference (e.g., Binz et al., 2025; de Varda et al., 2025). Here, we ask whether these systems exhibit modular organization resembling that found in human brains.

Beyond the well-established segregation between language and the rest of higher-level cognition (Fedorenko et al., 2024a,b; Mahowald, Ivanova et al., 2024), reasoning itself is divided into distinct systems—an organization that may support the robust, generalizable reasoning seen in humans. For example, social reasoning is selectively supported by a Theory of Mind network (Saxe & Kanwisher, 2003). Reasoning about the physical world draws on an Intuitive Physics network (Fischer et al., 2016). And formal reasoning, including mathematical and logical reasoning, draws on the so-called Multiple Demand network (Duncan, 2010; Fedorenko et al., 2013). When, how, and why this organization emerges remains debated. Is it beneficial for task performance, such that it will arise from any effective procedure for optimization of these tasks, whether biological evolution or deep learning? This question is hard to settle from brains alone, because biological tissue also faces metabolic pressures that have nothing to do with task performance. Neural activity has a high energy cost (roughly 20% of the human body's energy budget; Lennie, 2003), and representing information with a sparse code, in which only a small fraction of neurons is active at any time, saves a substantial amount of energy (Földiák & Young, 1995). Dedicating a separate set of neurons to each kind of task is one way to achieve such sparsity, so specialization could emerge as a solution to this energy budget. Evidence from artificial systems that face no metabolic pressure can therefore help us understand whether modularity can arise from optimizing task performance alone.

Modularity can emerge in systems whose underlying components are not themselves modular (Khona, Chandra, & Fiete, 2025). In standard neural network models, architectural modularity is not built in (cf. mixture-of-experts models, Cai et al., 2024; including brain-aligned experts, AlKhamissi et al., 2026a), but sometimes emerges spontaneously from task optimization (Csordás et al., 2020; Hod et al., 2021). For example, convolutional neural networks trained on both face and object recognition spontaneously segregate into two distinct subnetworks, one specialized for face recognition and one for object classification (Dobs et al., 2022). Simple RNN architectures trained on multiple tasks end up with subsets of neurons, each engaged only during one particular task (Yang et al., 2019). And some neurons in LLMs end up specializing for particular aspects of language processing (such as number agreement, Lakretz et al., 2019; Kryvosheieva et al., 2025), particular kinds of meanings (such as affective content, Radford et al., 2017; de Varda & Marelli, 2024), abstract reasoning (e.g., arithmetic, Stolfo et al., 2023; size comparison, Hanna et al., 2023), or knowledge retrieval (Dai et al., 2022; Meng et al., 2022). However, many of these were not grounded in the functional architecture of the human brain, and did not attempt to broadly characterize the systems supporting higher-level cognition in LLMs (cf. Yu et al., 2026).

Two recent studies have begun to bridge this gap. AlKhamissi et al. (2025) identified language-selective neurons in LLMs using an approach from neuroscience based on a 'localizer' contrast (Fedorenko et al., 2011) and demonstrated the causal role of those neurons in language tasks, but found mixed results when extending this approach to reasoning domains. AlKhamissi et al. (2026b) found neurons responding to contrasts targeting different forms of reasoning, but did not test for overlap or segregation between those sets of neurons, nor did they assess their causal role. Hanna et al. (2026) examined whether LLMs develop distinct circuits for *formal* linguistic tasks (e.g., those having to do with morphology, syntax, and word meanings) versus *functional* linguistic tasks (e.g., reasoning, knowledge retrieval), finding some degree of separation; but this broad formal–functional distinction (Mahowald, Ivanova et al., 2024) collapses many different cognitive capacities into a single 'functional' category, whereas human neuroscience has established that distinct systems support different forms of reasoning.

Here, we present a comprehensive, large-scale investigation of LLMs to test for the presence of human-like fine-grained modular organization—distinguishing not only language from reasoning, but also different reasoning domains from one another. In particular, we consider N=46 tasks spanning four cognitive domains: language processing, formal reasoning (including arithmetic and logical reasoning, and code comprehension), social reasoning, and physical reasoning. We then use circuit-level analyses to identify the neurons that are causally involved in each task and examine neuron overlaps across tasks. To foreshadow our key findings, the internal architecture of LLMs mirrors that of the human brain: tasks supported by the same brain network in humans recruit substantially overlapping circuits, while tasks supported by different networks rely on largely distinct sets of neurons.

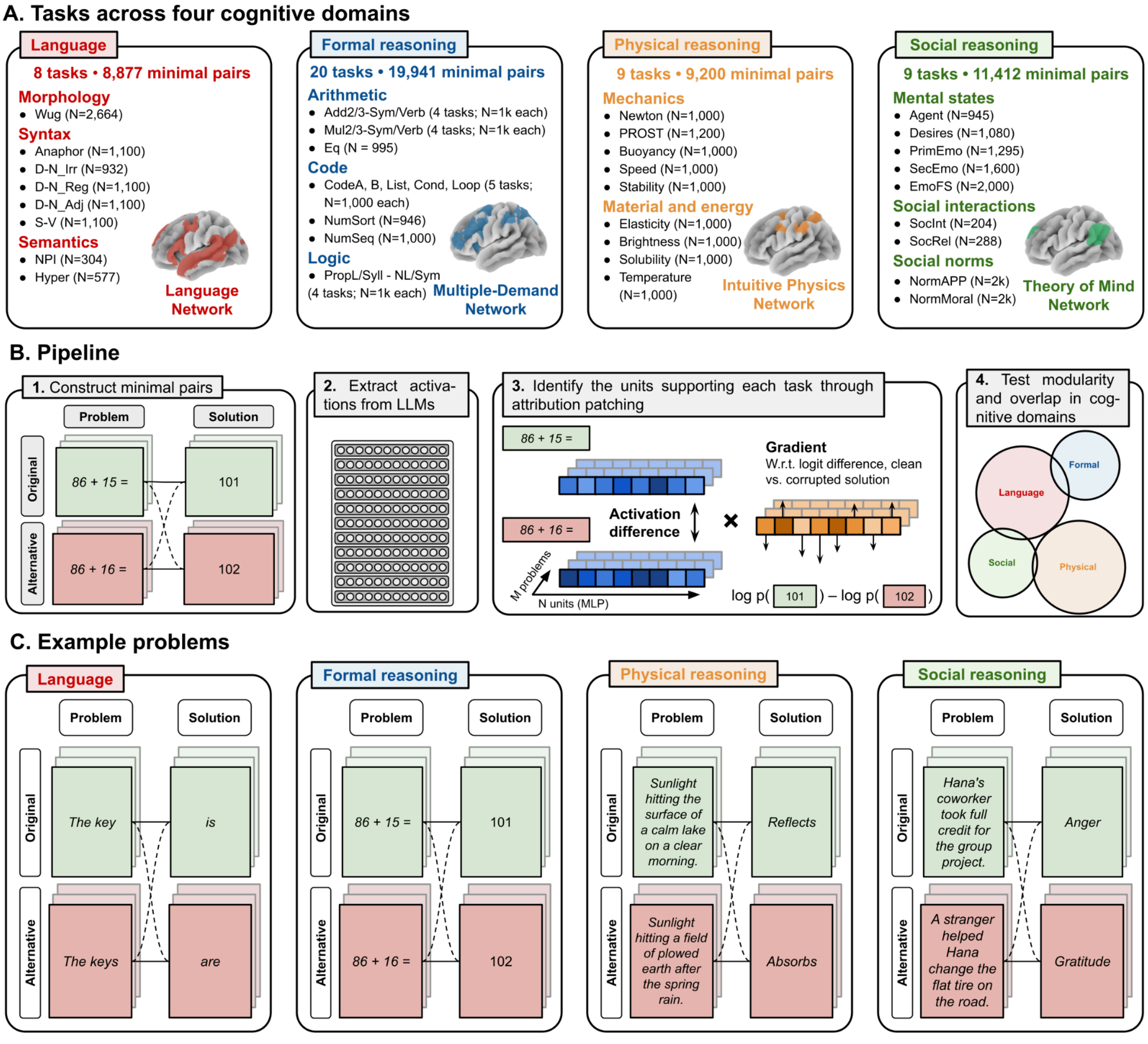

*Figure 1.* **Identifying domain-specific functional organization in large language models. (A)** Overview of 46 tasks organized into four cognitive domains, each grounded in a distinct functional brain network. N denotes the number of minimal pairs per task. **(B)** Pipeline for measuring modular organization across cognitive domains. (1) For each task, we construct minimal pairs of original and alternative inputs that elicit different correct continuations (here, two versions of an addition problem). (2) We run both inputs through the model and record activations and gradients at every multi-layer perceptron neuron. (3) Attribution patching identifies the relevance of each neuron with respect to a task by multiplying the activation difference between the original and alternative prompts by the gradient of the logit difference between the original and alternative continuations; this yields an estimate of each neuron's causal contribution to the task. This procedure linearly approximates the effects of replacing each neuron's activation with one from an alternative version of the problem. (4) We measure pairwise overlap of top-attributed neurons across tasks and cognitive domains. **(C)** Example minimal pairs for the four cognitive domains. For each task, the original and alternative inputs are minimally different but elicit different correct responses. In the language example (subject-verb agreement), the subject's grammatical number is changed (*the key is* / *the keys are*). In the formal reasoning example (two-digit addition), one operand is shifted by a small offset (86 + 15 / 86 + 16). In the physical reasoning example, a property of the depicted scene is altered, changing the relevant physical outcome (calm lake / plowed earth; reflects vs. absorbs sunlight). In the social reasoning example, the described event is altered, changing the inferred emotion (taking credit elicits anger; helping with a flat tire elicits gratitude). All tasks are operationalized as next-token prediction; many tasks include a task-specific prompt template before the input (see SI Section 2.5).

## Results

We use attribution patching (Syed et al., 2024), a gradient-based method that estimates the causal importance of all neurons simultaneously (Figure 1B; full methods in SI Section 1). For a given problem, it linearly approximates the effect of replacing each neuron's activation with one from an alternative version of the problem. We apply attribution patching across 46 tasks spanning four cognitive domains (Figure 1A), each corresponding to a distinct network in the human brain: language (Fedorenko et al., 2011, 2024a), multiple demand (Duncan, 2010; Assem et al., 2020), physical reasoning (Fischer et al., 2016), and theory of mind (Saxe & Kanwisher, 2003; see SI Section 2 for an overview of the tasks). Each task comprises a set of minimal pairs. Each pair consists of an original problem and an alternative version of the same problem that differ in some task-relevant dimension, such that the correct answer is reversed between the two variants (Figure 1B). Using these minimal pairs, attribution patching quantifies the contribution of individual neurons to task performance.

We then measured overlap among the top 0.1% most important neurons identified independently for each task and validated these results with targeted ablation experiments (see SI Section 5 for a replication across a range of attribution thresholds, top 0.05% to 5%). In these experiments, activations of task-critical neurons were replaced with alternative-input activations to assess their causal impact on performance across the full task battery. Analyses were conducted on six state-of-the-art language models ranging from 24B to 123B parameters (SI Section 8), all of which achieved strong performance on the task battery. Average performance was roughly balanced across domains (ranging from 81% in physical reasoning to 85% in the language tasks; SI Section 3).

**Modular organization of cognition in LLMs.** To quantify whether cognitive abilities of LLMs are supported by segregated systems, we computed the pairwise overlap in the top-0.1% neurons between all task pairs, averaged across six models. Indeed, LLMs show a modular functional organization that mirrors that of the human brain. First, tasks of the same type (e.g., arithmetic problems in different formats, or different forms of syntactic problems; Figure 2A) draw on the same neurons. Second, tasks of different types that are supported by the same network in the human brain (e.g., those involving math and computer code; Fedorenko et al., 2013; Ivanova et al., 2020) share substantially more neurons than tasks supported by different networks (e.g., math and physical reasoning). On average, within-domain overlap exceeded cross-domain overlap more than fourfold (12.9% vs. 3.0%, $p < 0.0001$). This pattern held within each of the four cognitive domains individually (SI Section 4.2) and in each of the six models individually (SI Section 4.1). The overlap structure itself was also highly consistent across models (mean pairwise Kendall's $\tau = 0.70 \pm 0.06$, all $p < 0.0001$). We further tested in a bottom-up way whether the modular structure we found in LLMs aligns with the domain boundaries defined by cognitive neuroscience, performing hierarchical clustering on the task-by-task overlap matrix and comparing the resulting groupings to our four predefined domains. The clustering successfully recovered the four predefined domains (adjusted Rand index = 0.78, $p < 0.0001$).

The modular organization was visible across network depth (Figure 2B). Although the four domain-relevant neurons were concentrated in the same mid-to-late layers, they relied on largely non-overlapping neurons within those layers; the Language neurons additionally showed a substantial presence in early layers. This pattern suggests a processing hierarchy in which linguistic computations parse the input from early layers, resulting in a representation of the problem on which downstream reasoning systems can operate. Importantly, this modular architecture did not emerge in a baseline model that did not achieve above-chance performance on the reasoning tasks in our battery (GPT-2; SI Section 7). This difference between high-performant models and the low-performant baseline model suggests that the modular architecture recovered by attribution patching is not an artifact of the datasets or the contrastive design; instead, it captures the underlying computations and is contingent on the model being able to actually solve the tasks.

The analysis of the low-performant GPT-2 model helps address another potential concern: namely, that the modular structure we observe might reflect nothing more than semantic differences between domains. After all, tasks in the same domain share vocabulary: physical reasoning tasks mention objects and materials, social reasoning tasks mention people and feelings. Perhaps a model that only represented the semantic content of the problems, without performing the relevant reasoning computations, would group the tasks the same way. However, despite being able to represent the meanings of social and physical words, GPT-2 recovers only the broad division between language and the rest of cognition, not the finer separation among the three reasoning domains (SI Section 7). Direct controls on the effects of semantic similarity in the prompts confirm that neuron overlap carries additional information about the computations of each domain independent of semantic similarity (SI Section 10).

**Modular causal effects on model behavior.** Attribution patching identifies which neurons should matter for a task, but it linearly approximates their causal role and does not directly test it. Thus, we performed ablation experiments across all pairs of tasks (2,070 pairs across 46 tasks excluding the diagonal). For each task pair, we identified the top 0.1% most important neurons for a "source" task (the pattern was consistent across threshold choices; SI Section 5.2) and then evaluated the model on a different "target" task with those neurons' activations replaced by their alternative-input values (similar results were obtained with mean- and zero-ablation; SI Section 6). If reasoning is modular, this intervention should impair target-task performance when source and target belong to the same domain ("within-domain" ablation) but not when they belong to different domains ("cross-domain" ablation). Within-domain ablations caused substantially larger accuracy drops than cross-domain ablations (mean ΔAccuracy averaged across models: within-domain 25.9% vs. cross-domain 2.5%; ratio 10.3×; $p < 0.0001$). This asymmetry holds for each domain individually, and in both directions of cross-domain ablation — a domain's neurons affect that domain's own tasks far more than other domains' tasks, and a domain's tasks are far more affected by its own neurons than by those of other domains (Fig. 2C; SI Section 4.2). Language neurons were the most selective, disrupting other Language tasks by 20.4% while affecting tasks in other domains by less than 0.5% (ratio 44.3×). Formal reasoning neurons showed the largest within-domain effect (32.0% vs. 3.0% cross-domain; ratio 10.6×). Physical reasoning neurons reduced within-domain accuracy by 16.0% (vs. 4.1% cross-domain; ratio 3.9×), and social reasoning neurons by 7.9% (vs. 2.0% cross-domain; ratio 4.0×). This pattern was highly consistent across all six individual models (mean pairwise Kendall's τ on the ablation effect matrix = 0.59 ± 0.05; 15 model pairs, asymptotic $p < 0.0001$). A qualitative inspection of the models' output after targeted ablations showed that lesioning neurons selectively required for performance on the language tasks largely preserved the models' reasoning abilities but introduced syntactic and morphological errors; conversely, lesioning physical reasoning neurons led the models to incorrect reasoning and conclusions but preserved the linguistic well-formedness of the output (see SI Section 9).

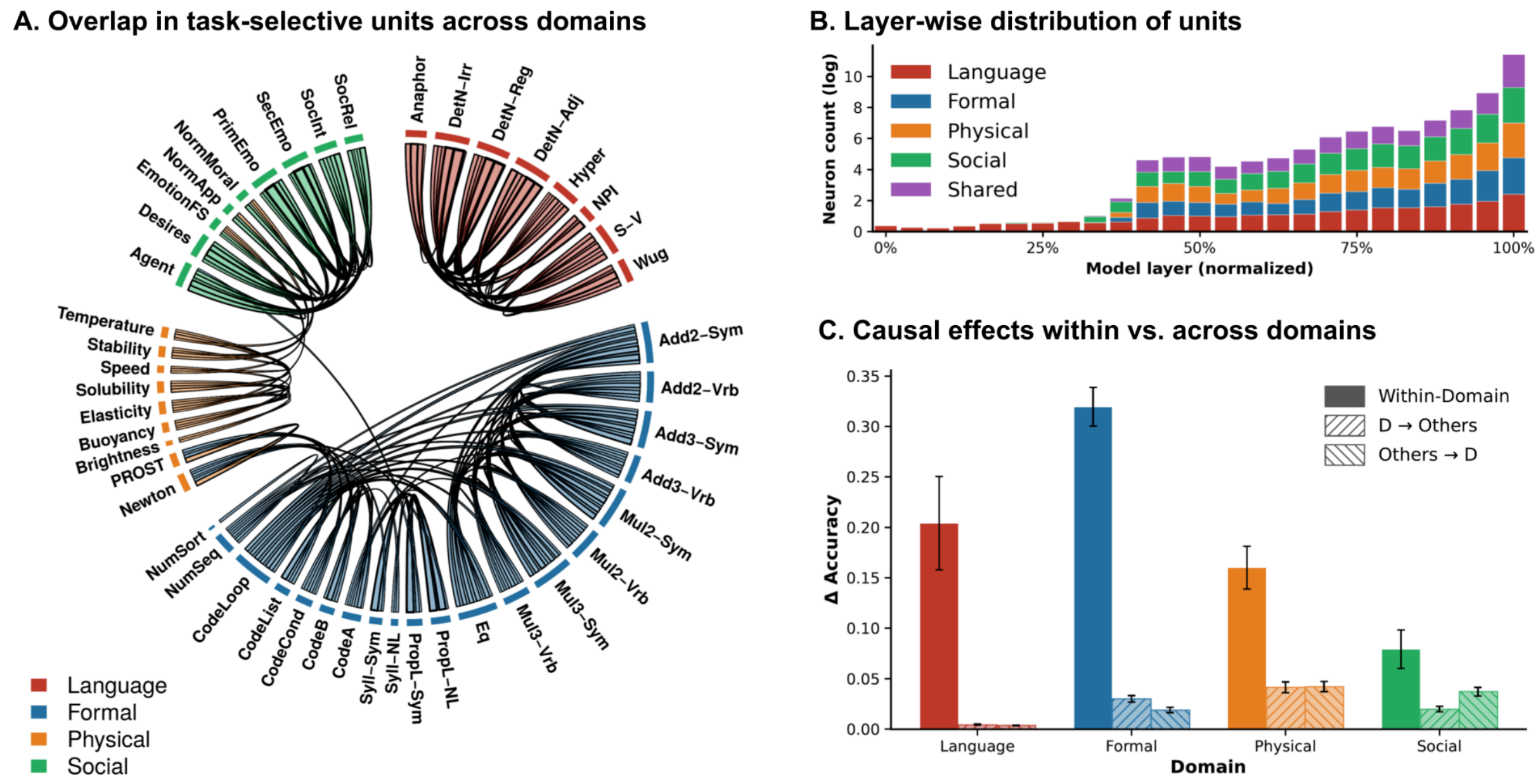

***Figure 2.*** **Large language models exhibit modular functional organization aligned with cognitive domains.** **(A)** Chord diagram of pairwise overlap between top 0.1% task-selective neurons across 46 tasks and 6 models. For each pair of tasks, we compute the Jaccard overlap of attributed neurons, apply doubly-stochastic normalization to remove domain-level scale differences, and average across models. Chords show pairs with normalized overlap above 5%, with thickness proportional to magnitude. Tasks are grouped by domain (color-coded): Language, Formal reasoning (arithmetic, logic, and code/algorithmic), Physical reasoning, and Social reasoning. Overlap is dense within domains and sparse between them. **(B)** Layer-wise distribution of top 0.1% attributed neurons, averaged across the six models and plotted on a log scale; layer indices are normalized to 0–100% to align models of different depths. Neurons attributed to two or more domains are classified as Shared. Domains coexist within the same layers using largely non-overlapping neurons. Only Language neurons are present in early layers. **(C)** Causal specificity of task-selective neurons. For each pair of source and target tasks, we ablated the top 0.1% neurons attributed to the source task and measured the accuracy drop on the target

task, then averaged across the six models. The solid bars indicate within-domain accuracy drops ($\Delta$ Accuracy over the baseline accuracy, which is roughly balanced across domains; SI Section 3), whereas the hatched bars indicate cross-domain accuracy drops. Error bars indicate SEM across task pairs. Within-domain ablations produce substantially larger drops than cross-domain ablations, indicating that task-level top neurons are causally involved in model performance almost exclusively for same-domain tasks.

## Discussion

Our results show that large language models—a new class of intelligent systems—develop the same modular organization that characterizes the human brain: language, formal reasoning, reasoning about the physical world, and social reasoning are supported by largely distinct sets of neurons; in contrast, tasks within a domain recruit substantially overlapping neurons. This convergence between brains and neural networks, in spite of the fact that the latter are shaped by an entirely different kind of optimization (gradient descent on next-token prediction), suggests that modularity may constitute a general principle of how intelligent systems tend to be organized.

These findings also offer leverage on a question that has long been difficult to address by looking at the brain alone: *why* such organization arises in the first place. Several explanations for modularity have been proposed in cognitive neuroscience (for an overview, see Kanwisher, 2010). One influential idea appeals to metabolic constraints that are specific to biological tissue: a brain that activates only a small fraction of neurons for any given task uses less energy than one in which activity is broadly distributed (Földiák & Young, 1995). This pressure does not apply to the neural networks we studied here since a transformer's forward pass has no variation in metabolic cost in any biologically meaningful sense: nothing in the loss function penalizes the number of neurons that are active on a given input. That a recognizably human-like modular organization nonetheless emerges in LLMs suggests that this biological pressure, whatever role it plays in shaping the functional infrastructure of the human cortex, is not necessary for functional specialization to emerge.

If not for metabolic constraints, why then may LLMs develop a modular organization? To approach this question, it is useful to think about the problem that LLMs are trained to solve (McCoy et al., 2024). Transformer-based LLMs are trained to predict the next token across enormous, heterogeneous corpora, in which language parsing and different forms of reasoning often need to interact in supporting the prediction. Consider the complex task of predicting the next word towards the end of a detective novel: this task can simultaneously require deductive inference to narrow down a set of suspects, social reasoning to attribute motives, and physical reasoning to understand the mechanics of the crime (e.g., whether a bullet trajectory through a window is geometrically possible). And all these reasoning computations need to interact with the linguistic processes needed to parse sentence structures.

When several forms of reasoning must be engaged simultaneously over the same input, the system faces a pressure to keep these computations from interfering with one another. This pressure operates at two different levels. The first concerns processing: representing several kinds of information simultaneously is feasible only if the codes for each remain separable, because intermixed codes would collide and degrade the representations (Reddy & Kanwisher, 2007; Kanwisher, 2010). The second concerns learning: when the neurons responsible for different computations are intermixed, updates that improve performance in one domain will tend to perturb representations relevant to another, producing the kind of interference that has long been recognized as a central obstacle to learning in distributed systems (McCloskey & Cohen, 1989; French, 1999). Allocating distinct subpopulations of neurons to distinct kinds of computation offers protection from both types of interference: it allows the system to represent the relevant information separately, and to acquire new information about one domain without overwriting what it has learned about another. The convergent emergence of modularity in biological brains and in LLMs suggests that this pressure may, on its own, suffice to give rise to modularity in intelligent systems that have to solve diverse types of problems.

Our findings also have practical implications. For cognitive neuroscience, our approach may be used to generate predictions about which brain networks a novel task will engage in humans, allowing to derive hypotheses from LLMs to be tested with fMRI. Moreover, LLMs offer a way to study something that is very hard to investigate in the brain: how different systems transmit information to one another. Tracking information as it flows from one network to another requires both high spatial and high temporal resolution—a combination that is hard to achieve in human neuroimaging. In an LLM, in contrast, we have simultaneous access to the full set of neurons and their connections, so we can develop and test mechanistic hypotheses about how, say, the language system feeds input to a reasoning system.

For artificial intelligence research, future work may clarify the relationship between modular organization and task performance, potentially informing how future systems are designed. This question is especially relevant given the recent surge of mixture-of-experts models (Cai et al., 2024). Their adoption has been motivated largely by efficiency, since routing each token to a subset of experts reduces the number of active parameters (Shazeer et al., 2017), but our findings raise the possibility that modular organization carries additional computational advantages. Recent work points in this direction, showing that inducing brain-like specialization across experts preserves reasoning performance and at the same time makes models more interpretable and steerable (AlKhamissi et al., 2026a).

More broadly, our results illustrate the value of LLMs as a second class of intelligent systems against which to test claims about the structure of cognition and the general constraints on intelligence. To the extent that a feature of human cognition emerges in systems shaped by different pressures, we have reason to suspect that it reflects a general principle of intelligent organization. Modularity, our findings suggest, is one such feature; whether others (e.g., the existence of grid-like codes for representing abstract conceptual spaces; Constantinescu et al., 2016; the existence of critical or sensitive periods during which specific abilities are best acquired; Knudsen, 2004) follow the same logic is an open question that can be addressed by this kind of comparative approach.

**Acknowledgements:** We are grateful to Nancy Kanwisher for conceptual input throughout the project and helpful feedback on the draft. JA was supported by research funds from the MIT Siegel Family Quest for Intelligence. EF was supported by research funds from the McGovern Institute for Brain Research, the Simons Center for the Social Brain, and the MIT Siegel Family Quest for Intelligence. AGdV was supported by the K. Lisa Yang ICoN Center Postdoctoral Fellowship. JA and EF and this research were partially supported by the Defense Advanced Research Projects Agency (DARPA) AIQ program through the DARPA CMO contract number HR00112520025.

**Contributions:** **PH:** Conceptualization, Methodology, Investigation, Data curation, Visualization, Formal analysis, Validation, Software, Writing-Original draft, Writing-Review and editing. **JA:** Conceptualization, Methodology, Writing-Review and editing, Supervision. **EF:** Conceptualization, Methodology, Writing-Original draft, Writing-Review and editing, Supervision, Project administration. **AGdV:** Conceptualization, Methodology, Investigation, Data curation, Visualization, Formal analysis, Validation, Writing-Original draft, Writing-Review and editing.

# Supplementary Information

## 1. Methods

### 1.1 Datasets

We considered a total of 46 tasks broadly covering four cognitive domains, each corresponding to a well-characterized functional network in the human brain: language (Fedorenko et al., 2011, 2024a), multiple demand reasoning (Duncan, 2010; Fedorenko et al., 2013), social or 'Theory of Mind' reasoning (Saxe & Kanwisher, 2003), and physical reasoning (Fischer et al., 2016). For each task, we constructed pairs of 'original' (e.g., 65 + 16 =) and minimally different 'alternative' (86 + 16 =) inputs in which the problem is modified so that a different answer becomes correct (101 vs. 102) while surface features (length, format, and general complexity) are held constant. We then localize the neurons most critical for each task by measuring, for each neuron, how much restoring its original activation into an alternative forward pass recovers the model's correct behavior (SI Section 1.3). Within each task, we filtered our problem set to retain only items where the model answered both the original and alternative versions correctly, requiring a both-correct rate of at least 60% per task (with chance level being 25%). Full task specifications, example items, and item counts are provided in SI Section 2.

*Language.* We included 8 tasks covering morphology, syntax, and lexical semantics. Syntactic tasks were drawn from BLiMP (Warstadt et al., 2020) and included subject-verb agreement, determiner-noun agreement (regular, irregular, and with intervening adjectives), anaphor gender agreement, and negative polarity item licensing. Morphological generalization was tested with a Wug-style task (Hanna et al., 2026; Keuleers & Brysbaert, 2010). Lexical semantics was tested with a hypernymy task adapted from Hanna et al. (2026). Alternative inputs were constructed by modifying the relevant cue so that the opposite continuation became correct (e.g., original: "Some customers know these ___" → "men"; alternative: "...know this ___" → "man").

*Multiple demand reasoning.* We included 20 tasks spanning arithmetic (9), logic (4), and code and algorithms (7). Arithmetic tasks tested two- and three-operand addition/subtraction and multiplication/division in both symbolic ('132 - 129 =') and verbal formats ('*One hundred thirty-two minus one hundred twenty-nine equals*'), plus linear equations. Logic tasks tested syllogistic and propositional reasoning in both natural-language and symbolic formats. Code and algorithm tasks included Python output prediction for functions involving arithmetic, variable assignment, conditionals, list indexing, and loops, as well as number sorting and number sequence completion. Number sorting and sequence completion were adapted from Reasoning Gym (Stojanovski et al., 2025); the remaining tasks were constructed by us. Details on dataset construction are provided in SI Section 2.2. Alternative inputs were created by modifying a single operand, operator, quantifier, or control-flow element so that a different answer became correct.

*Physical reasoning.* We included 9 tasks probing intuitive physics. Two tasks (Newton, PROST) were adapted from existing benchmarks (Wang et al., 2023; Aroca-Ouellette et al., 2021). Seven additional tasks probed specific physical properties (buoyancy, elasticity, solubility, brightness, speed, stability, temperature) and were constructed by us. Detailed dataset construction is provided in SI Section 2.3. Alternative inputs were created by modifying the physical context (e.g., material, mass, configuration) so that the opposite outcome became correct.

*Social reasoning.* We included 9 tasks spanning emotion inference (3), belief and goal attribution (2), social interactions and relations (2), and norm judgment (2). Emotion tasks tested the inference of primary emotions (e.g., joy after winning), secondary emotions (e.g., pride after public praise), and complex emotions in multi-sentence scenarios. Belief and goal attribution tasks tested whether the model could infer an agent's beliefs from evidence and goals from behavior. Social interaction and relation tasks tested inference of interpersonal dynamics and social roles. Norm judgment tasks tested moral judgments and assessments of social appropriateness. Belief attribution and social relations were adapted from EWOK (Ivanova, Sathe, Lipkin, et al., 2025); the remaining tasks were constructed by us. Detailed dataset construction is provided in SI Section 2.4. Alternative inputs were constructed by modifying the social context so that the opposite mental state, role, or judgment became correct (e.g., original: "Ali just won the competition. How does Ali feel?" → "joyful"; alternative: "Ali just lost…" → "sad").

We assigned each task to a domain based on the characterization of the functional network that supports each domain in the human brain. We note, however, that many of the tasks we include have not been directly tested against neural responses. For some tasks the mapping rests on the broad characterization of the relevant network (e.g., the Theory of Mind network for social reasoning tasks). Our domain assignments therefore reflect well-motivated hypotheses about which network each would recruit.

### 1.2 Models

We considered six autoregressive, instruction-tuned LLMs spanning from 24 to 123 billion parameters and four model families: Mistral-Small-24B-Instruct (Mistral AI, 2025), Qwen2.5-32B-Instruct (Yang et al., 2024), OLMo-2-32B-Instruct (OLMo et al., 2024), Llama-3.1-70B-Instruct (Grattafiori et al., 2024), Qwen2.5-72B-Instruct (Yang et al., 2024), and Mistral-Large-123B-Instruct (Mistral AI, 2024). All models have publicly available weights. This selection spans four distinct model families and a roughly 5-fold range in parameter count, allowing us to test whether the modular organization we observe is consistent across architectures and scales. We used instruction-tuned variants to ensure reliable task performance across the diverse reasoning domains in our benchmark; prior work has shown that the underlying circuit structure is largely preserved between base and instruction-tuned models (Hanna et al., 2026; Kissane et al., 2024; Prakash et al., 2024). Model architectural details (number of layers, attention heads, and MLP dimensions) are provided in SI Section 8.

### 1.3 Circuit localization via attribution patching

To identify the neurons that support each task, we use Attribution Patching (Syed et al., 2024), a gradient-based method that efficiently estimates the causal importance of each neuron in the model. Attribution patching can be applied at different levels (attention heads, layers, individual neurons, edges); in this work, we focus on neurons, i.e., the hidden neurons in each transformer layer's feed-forward intermediate layer. Attribution patching achieves near state-of-the-art performance on the Mechanistic Interpretability Benchmark (Mueller et al., 2025), confirming that it approximates well the results of more expensive causal interventions.

Intuitively, attribution patching identifies neurons that satisfy two criteria: (1) their activations differ between original and alternative inputs, and (2) the model's prediction is sensitive to these differences. Concretely, for each MLP neuron $i$, we compute an attribution score:

$$attribution_i = (z_i^{original} - z_i^{alternative}) \nabla_i \mathcal{L}$$

where $z_i^{original}$ and $z_i^{alternative}$ are the neuron's activations on the original and alternative inputs, and $\nabla_i \mathcal{L}$ is the gradient of the task metric with respect to the neuron's activation. The activation difference captures how much the neuron responds to the relevant contrast, while the gradient captures how much the model's output depends on the neuron. Neurons that score high on both are causally important for the task. As an illustration, consider the subject-verb agreement task: the original input "*The keys on the table* ___" requires *are*, while the alternative input "*The key on the table* ___" requires *is*. A neuron that activates strongly for plural subjects and weakly for singular subjects will have a large activation difference: it sensitizes to the contrast between "keys" and "key". But detecting the contrast is not enough: many neurons may respond differently to the two nouns without affecting the choice of the verb. The gradient term filters for relevance in the model's output: it measures how much the model's prediction shifts when the neuron's activation is perturbed. A neuron scores highly only if it both detects the number contrast (large activation difference) and occupies a position in the computation where that signal propagates forward to change the model's output (large gradient). We compute attribution scores at the input to each layer's MLP down-projection, evaluated at the final prompt token position (immediately preceding the answer), and sum scores across all examples within a task. To ensure that activation differences are meaningful at each token position, we verified that original and alternative prompts within each example pair have identical token sequence lengths; examples failing this constraint were excluded.

### 1.4 Overlap analysis

After identifying the top-attribution neurons for each task independently, we measured the degree of separability in the neurons supporting each task by computing pairwise neuron overlap. Within each model, we focused on neurons with positive attribution scores. These are neurons whose original activations actively support the correct task behavior. For each task, we selected the top 0.1% of such neurons (from 1311 to 2523 neurons across models). For each pair of tasks (A, B), we computed the Jaccard index as the size of the intersection of their top-neuron sets divided by the size of their union. This yielded a symmetric task-by-task overlap matrix for each model. To aggregate across the six models, we took the union of tasks covered by at least one model, yielding a 46-task set. For each task pair (A, B), we averaged the Jaccard values from all models in which both A and B were present ( i.e., if 3 models had both tasks, we summed their 3 Jaccard values and divided by 3). For the chord diagram in Figure 2A, we additionally apply doubly-stochastic normalization to the overlap matrix so that each row and column sums to one, which improves readability by equalizing tasks with broadly versus narrowly distributed attribution profiles. This normalization is used for visualization only; all quantitative analyses and statistical tests operate on raw overlap values.

We assessed the modular structure of the overlap matrix using three complementary statistical analyses. First, to test whether within-domain overlap is significantly higher than cross-domain overlap, we performed a permutation test: we computed the observed difference between the mean within-domain overlap (excluding the diagonal) and the mean cross-domain overlap, then re-computed this difference under 10,000 random permutations of the domain labels assigned to tasks. Second, to test whether unsupervised clustering of the overlap matrix recovers the four cognitive domains, we constructed a $k$-nearest-neighbor graph ($k$=5) from the overlap matrix, performed hierarchical clustering with average linkage, cut the dendrogram into four clusters, and computed the Adjusted Rand Index (a measure of clustering agreement corrected for chance level) between the resulting cluster labels and the ground-truth domain labels. Third, to test whether the modular organization is consistent across models, we computed pairwise Kendall's rank correlation ($\tau$) between the upper triangles of every pair of model-specific overlap matrices, restricted to tasks present in both models.

**1.5 Causal intervention**

To validate that the neurons identified by attribution patching are causally important for task performance, we performed targeted ablation experiments. Rather than zeroing out neuron activations (a common but problematic approach, since zero values lie outside the distribution of natural model activations and can produce performance drops unrelated to the actual function of the ablated neurons'; Hanna et al., 2026), we replaced each neuron's original activation with its corresponding activation from the alternative forward pass. This intervention keeps activations within the model's natural distribution but is expected to change the answer of the model, if the neuron is causally involved in the task.

Crucially, our ablation tests the neurons' impact on performance in tasks different from those used to identify the critical neurons. For each pair of tasks (source ≠ target), we asked: are neurons identified as critical for the source task also causally involved in the target task? To answer this, we first identified the top 0.1% of neurons based on the source task's attribution scores. We then ablated these neurons while the model performed the target task. Specifically, we ran the target task's original input through the model but replaced the activations of the selected 0.1% neurons with the values they would have taken on the target task's alternative input, thus injecting alternative information into precisely those neurons that the source task depends on. We measured the resulting drop in target-task accuracy relative to the unperturbed original baseline. If neurons critical for one task also disrupt a different task, this suggests the two tasks rely on (at least partially) shared resources. We excluded the diagonal (ablating a task's own neurons and evaluating on the same task), as this would trivially show an effect and would not test whether the neurons' causal contribution is higher within vs. across cognitive domains.

If reasoning is modularly organized in the model, ablating source-task neurons should impair performance on target tasks within the *same* domain substantially more than on target tasks from *different* domains. This logic mirrors evidence from neuropsychology, where damage to a brain region selectively impairs functions supported by that region while leaving other functions intact (e.g., damage to the left-hemisphere language regions causing profound impairments in linguistic abilities while preserving formal reasoning; Kean et al., 2025). We quantified ablation effects as the drop in target-task accuracy relative to the unperturbed original baseline, and compared the distribution of within-domain ablation drops to cross-domain ablation drops using the same statistical analyses used for testing overlap.

## 2. Data description

### 2.1 Language datasets

These tasks include tests for morphological, syntactic, and lexical semantic processing, the operations supported by the language network in the human brain (Fedorenko et al., 2011, 2024a). Six tasks (Anaphor, D-N_Irr, D-N_Reg, D-N_Adj, NPI, S-V) are drawn from BLiMP (Warstadt et al., 2020), and two (Hyperny, Wug) are drawn from Hanna et al. (2026). All tasks are constructed as minimal pairs in which the original and alternative inputs differ in a single syntactic or lexical feature that flips the correct continuation.

**Anaphor (N = 1,100)**. The anaphor agreement task tests whether the model resolves anaphoric pronouns based on the gender of the antecedent. The corruption swaps the antecedent to a different-gender name, flipping the correct pronoun. Original: “Katherine can’t help herself.” Alternative: “Thomas can’t help herself.” (correct continuation flips from herself to himself).

**D-N_Irr (N = 932)**. The determiner–noun agreement (irregular) task tests number agreement between a demonstrative and an irregular plural noun. The corruption swaps the determiner, flipping singular/plural expectation. Original: “Grace noticed that axis.” Alternative: “Grace noticed those axes.” (correct continuation flips from axis to axes).

**D-N_Reg (N = 1,100)**. The determiner–noun agreement (regular) task is the same probe with regular plurals. The corruption swaps the determiner. Original: “This hat disgusts that boy.” Alternative: “This hat disgusts those boys.” (correct continuation flips from boy to boys).

**D-N_Adj (N = 1,100).** The determiner–noun agreement task with an intervening adjective tests whether number agreement is preserved across the adjective. The corruption swaps the determiner. Original: “All guests had broken this bad dish.” Alternative: “All guests had broken these bad dishes.” (correct continuation flips from dish to dishes).

**Hyper (N = 577).** The hypernymy task probes category knowledge: given a list of category members, the model must continue with the category label. The corruption substitutes members of a different category, flipping the correct label. Original: “sapphires, and other gems.” Alternative: “sycamores, and other trees.” (correct continuation flips from gems to trees).

**NPI** (N = 304). The negative polarity item task tests licensing of NPIs such as *‘any’*, which require a negative licensor. The corruption removes the matrix-clause negation, making *‘some’* the correct continuation. Original: “No journalists that contacted no surgeon have broken any.” Alternative: “The journalists that contacted no surgeon have broken some.” (correct continuation flips from any to some)

**S-V (N = 1,100).** The subject–verb agreement task tests number agreement between subject and verb. The corruption flips the subject's number, flipping the correct verb form. Original: "The actresses stand." Alternative: "The actress stands." (correct continuation flips from stand to stands)

**Wug (N = 2,664).** The wug task tests morphological generalization to nonce words, requiring the model to apply regular plural/singular morphology to novel forms. The corruption swaps the singular/plural context. Original: "First, there were two utmesitions. Now there is one utmesition." Alternative: "First, there was one utmesition. Now there are two utmesitions." (correct continuation flips from ition to itions)

## 2.2 Formal reasoning datasets

The formal reasoning datasets span arithmetic, propositional and syllogistic logic, code execution, and simple algorithmic tasks. They target the multiple-demand network, which in the human brain supports domain-general executive control and is engaged across many cognitively demanding tasks involving formal reasoning, including logic, mathematical reasoning (Fedorenko et al., 2013), and computer code comprehension (Ivanova et al., 2020). Each task is presented as a single prompt whose correct continuation is fully determined by the rule; the corruption modifies one operand, operator, quantifier, comparison, or directive so that the correct continuation flips to a different token. Three tasks (Eq, NumSeq, NumSort) are taken from Reasoning Gym (Stojanovski et al., 2025). The rest are generated by us.

**Add2-Sym (N = 1,000).** The two-operand symbolic addition/subtraction task tests integer arithmetic in standard symbolic form. The corruption changes one operand, flipping the result. Original: "132 − 129 =" Alternative: "130 − 129 =" (correct continuation flips from 3 to 1)

**Add2-Vrb (N = 1,000).** The two-operand verbal addition/subtraction task is the same probe in fully spelled-out English. The corruption changes one operand. Original: "one hundred and sixteen plus six hundred and ninety-five equals" Alternative: "two hundred and sixteen plus six hundred and ninety-five equals" (correct continuation flips from eight hundred and eleven to nine hundred and eleven)

**Add3-Sym (N = 1,000).** The three-operand symbolic addition/subtraction task extends Add2-Sym to chained operations. The corruption changes one operand. Original: "149 + 577 + 205 =" Alternative: "149 + 574 + 205 =" (correct continuation flips from 931 to 928)

**Add3-Vrb (N = 1,000).** The three-operand verbal addition/subtraction task is the verbal version of Add3-Sym. The corruption changes one operand. Original: "three hundred and sixty-eight plus two hundred and forty-seven minus five hundred and fourteen equals" Alternative: "… minus five hundred and sixteen equals" (correct continuation flips from one hundred and one to ninety-nine)

**Mul2-Sym (N = 1,000).** The two-operand symbolic multiplication/division task tests multiplicative arithmetic in symbolic form. The corruption changes one operand. Original: "666 / 18 =" Alternative: "486 / 18 =" (correct continuation flips from 37 to 27)

**Mul2-Vrb (N = 1,000).** The two-operand verbal multiplication/division task is an adaptation of Mul2-Sym in verbal format. The corruption changes one operand. Original: "nine hundred and twenty-eight divided by thirty-two equals" Alternative: "eight hundred and sixty-four divided by thirty-two equals" (correct continuation flips from twenty-nine to twenty-seven)

**Mul3-Sym (N = 1,000).** The three-operand symbolic multiplication/division task extends Mul2-Sym to chained operations. The corruption changes one operand. Original: "25920 / 18 / 96 =" Alternative: "25920 / 15 / 96 =" (correct continuation flips from 15 to 18)

**Mul3-Vrb (N = 1,000).** The three-operand verbal multiplication/division task is the verbal version of Mul3-Sym. The corruption changes one operand, flipping the result. Original: "nine thousand and one hundred divided by ten divided by sixty-five equals" Alternative: "nine thousand and one hundred divided by twenty divided by sixty-five equals" (correct continuation flips from fourteen to seven)

**Eq (N = 995).** The simple equation task requires the model to solve a one-variable linear equation. The corruption changes the constant term, flipping the solution. Original: "Solve for u: 19*u + 14 = 413" Alternative: "Solve for u: 19*u + 14 = 337" (correct continuation flips from 21 to 17)

**Syll-NL (N = 1,000).** The natural-language syllogism task tests classical syllogistic reasoning over quantified statements. The corruption swaps a quantifier (Some ↔ All), flipping the validity judgment. Original: "All chefs are doctors. Some writers are not doctors. Does it follow that some writers are not chefs?" Alternative: "… Does it follow that all writers are not chefs?" (correct continuation flips from Yes to No)

**Syll-Sym (N = 1,000).** The symbolic syllogism task is the same probe rendered in first-order logic notation (∀, ∃, →). The corruption swaps the question polarity (right ↔ wrong). Original: "∀x(G(x)→L(x)). ∀x(G(x)→H(x)). Which is right? ∃x(H(x)∧L(x)) or ∀x(H(x)→¬L(x))?" Alternative: "… Which is wrong?" (correct continuation flips from ∃x(H(x)∧L(x)) to ∀x(H(x)→¬L(x)))

**PropL-NL (N = 1,000).** The natural-language propositional logic task tests multi-step inference over conditional chains. The corruption swaps the question polarity (right ↔ wrong). Original: "If farmers are proud, then smart. If smart, then bold. Farmers are proud. Which is right?" Alternative: "… Which is wrong?" (correct continuation flips from 'farmers may not be bold' to 'farmers are bold')

**PropL-Sym (N = 1,000).** The symbolic propositional logic task is the same probe in propositional notation (∨, ¬, →). The corruption swaps the question polarity. Original: "U ∨ W. ¬U. Which is right? ¬W or W?" Alternative: "… Which is wrong?" (correct continuation flips from W to ¬W)

**CodeA (N = 1,000).** The code-A task is a mixed block spanning five Python control-flow patterns in roughly equal proportions (~200 examples each): multi-step variable tracing, conditional branching, loop accumulation, function calls with operator manipulation, and modulo checks. The model predicts the printed output of a short function, and the corruption changes a single token, typically an operator, so the output flips. Representative example (variable-tracing sub-pattern). Original: def solve(x): y=x*4; z=y+3; return z; print(solve(4)) Alternative: … z=y*3 … (correct continuation flips from 19 to 48).

**CodeB (N = 1,000).** The code-B task is a mixed block spanning five data-structure and string-operation patterns in roughly equal proportions (~200 examples each): list indexing, split-and-index on strings, dictionary lookup, string concatenation order, and variable swap. As in code-A, the model predicts the printed output and the corruption flips a single token. Representative example (variable-swap sub-pattern). Original: def pick(): x= 'open'; y=

‘shut’; x,y=y,x; return y Alternative: … return x (correct continuation flips from open to shut).

**CodeCond (N = 1,000).** The code-conditional task is an isolated single-pattern dataset: every example is one if/else block whose comparison selects between two return values. The corruption flips the comparison operator (>= → <=). Original: if score >= 3: return ‘win’ Alternative: if score <= 3: return ‘win’ (correct continuation flips from win to lose).

**CodeList (N = 1,000).** The code-list task is an isolated single-pattern dataset focusing on list indexing. The corruption changes the integer index. Original: options=[‘mango’, ‘banana’, ‘plum’, ‘cherry’, ‘apple’]; return options[4] Alternative: … return options[1] (correct continuation flips from apple to banana).

**CodeLoop (N = 1,000).** The code-loop task is an isolated single-pattern dataset focusing on loop-based accumulation. The corruption changes the range upper bound. Original: for n in range(5): x*=(n+1); return x Alternative: for n in range(2): … (correct continuation flips from 120 to 2).

**NumSort (N = 946).** The number-sequence task asks the model to fill in a missing element of an arithmetic sequence. The corruption moves the missing position, flipping the answer. Original: “9, ?, 2, 1, 0, 0” Alternative: “9, 4, 2, 1, 0, ?” (correct continuation flips from 4 to 0)

**NumSeq (N = 1,000).** The number-sorting task asks the model to sort a list of numbers in ascending or descending order. The corruption swaps the directive (ascending ↔ descending), reversing the target list. Original: “Sort in ascending order: −43.7, 7.1, 1.2, 11.6” Alternative: “Sort in descending order: …” (correct continuation flips from [−43.7, 1.2, 7.1, 11.6] to [11.6, 7.1, 1.2, −43.7])

## 2.3 Physical reasoning datasets

The physical reasoning datasets test intuitive physics, predicting how everyday objects behave under gravity, contact, light, heat, and fluid interactions. They target the physical-reasoning network identified in human neuroimaging studies (Fischer et al., 2016). Two of the tasks (Newton, PROST) are adapted from prior physical-commonsense benchmarks (Wang et al., 2023; Aroca-Ouellette et al., 2021). The remaining seven are newly generated property-specific tasks generated from templates. The templates were created by Claude Opus 4.5 and manually verified by the authors. The resulting tasks share a two-shot in-context template: each prompt presents two labeled exemplar scenarios followed by a third target scenario, and the model must continue with the property label (e.g., floats / sinks). The few-shot examples were included because model accuracy was too low in a zero-shot format. For these seven tasks, the corruption replaces the target scenario with one whose physical outcome flips, while keeping the two exemplars fixed. An example is shown below:

**Prompt (shared across Original and Alternative):**

What is the physical outcome in the following contexts?

Context: Sunlight hitting a stack of wet peat moss in the garden bed. Answer: absorbs

Context: A lamp shining on a smooth glass window pane from inside at night. Answer: reflects

Context: [TARGET] Answer: ___

**Original target:** A flashlight pointed at a piece of rough charcoal on the table. (→ absorbs)

**Alternative target:** A flashlight pointed at the shiny steel blade of a kitchen knife. (→ reflects)

For compactness, the remaining task examples below show only the target scenarios (original / alternative);

**Newton (N = 1,000).** The Newton task (Wang et al., 2023) tests intuitive physical commonsense about everyday object properties such as weight, fragility, sharpness, size, and suitability for a given context. Each example presents two objects and asks which is appropriate for some everyday situation; the corruption swaps the question word (e.g., top ↔ bottom, keep ↔ remove), flipping which of the two objects is the correct answer. Original: "I am packing a backpack. Which of clothes hamper and ice cream should I put at the top?" Alternative: "… should I put at the bottom?" (correct continuation flips from ice cream to clothes hamper)

**PROST (N = 1,200).** The PROST task (Aroca-Ouellette et al., 2021) tests physical commonsense about everyday object properties such as stackability, friction, fragility, weight, bounciness, and similar affordances. Each example presents two objects (or two surfaces) in some everyday scenario and asks which one better fits the comparative; the corruption swaps the comparative (e.g., easier ↔ harder, more likely ↔ less likely), flipping which of the two is the correct answer. Original: "A person is trying to stack a bottle and a brick. Which one is easier to stack?" Alternative: "… Which one is harder to stack?" (correct continuation flips from brick to bottle)

**Brightness (N = 1,000).** The brightness task tests whether a surface absorbs or reflects incident light. The corruption replaces the target scenario with one of opposite optical behavior. For example, Original: "A flashlight pointed at a piece of rough charcoal on the table." Alternative: "A flashlight pointed at the shiny steel blade of a kitchen knife." (correct continuation flips from absorbs to reflects)

**Buoyancy (N = 1,000).** The buoyancy task tests whether an object floats or sinks in a given liquid. The corruption replaces the target scenario with one of opposite buoyant behavior. Original: "Ravi dropped a hollow rubber duck into the bath tub water." Alternative: "Ravi dropped a heavy iron chain into the deep swimming pool." (correct continuation flips from floats to sinks)

**Elasticity (N = 1,000)**. The elasticity task tests whether an object bounces or breaks on impact. The corruption replaces the target scenario with one of opposite elastic behavior. For example: Original: "A glass lamp shade that fell from the counter onto the tile floor." Alternative: "A racquetball Luca served hard against the wall of the court." (correct continuation flips from breaks to bounces)

**Solubility (N = 1,000).** The solubility task tests whether a solid dissolves or settles when placed into a liquid. The corruption replaces the target scenario with one of opposite solubility behavior. Target — Original: "A spoonful of ground pottery Noor stirred into a cup of still water." Alternative: "A piece of hard toffee Noor dropped into a cup of boiling water." (correct continuation flips from settles to dissolves)

**Speed (N = 1,000).** The speed task tests whether motion in a given setting is fast or slow given friction, slope, or driving force. The corruption replaces the target scenario with one of opposite predicted speed. For example: Original: "A block is moving on a surface covered in a layer of tar." Alternative: "A rock is tumbling down a steep bare cliff with no trees." (correct continuation flips from slow to fast)

**Stability (N = 1,000).** The stability task tests whether a configuration of objects is stable or unstable under gravity. The corruption replaces the target scenario with one of opposite stability. For example: Original: "Leo put a heavy bowling ball on a tall thin plastic stand." Alternative: "Leo put a heavy glass paper weight flat on the office desk." (correct continuation flips from unstable to stable).

**Temperature (N = 1,000).** The temperature task tests whether an object will melt or freeze given the surrounding thermal environment. The corruption replaces the target scenario with one of opposite thermal outcome. Target — Original: "A bar of soap left on the edge of a tub full of steaming water." Alternative: "A cup of water left in the garage with no heater on a subzero night." (correct continuation flips from melt to freeze)

## 2.4 Social reasoning datasets

The social reasoning datasets test mental-state attribution and social judgment such as beliefs, desires, emotions, social roles, and norm-conformity. They target the theory-of-mind (ToM) network identified in human neuroimaging studies (Saxe & Kanwisher, 2003). Most tasks present a one-sentence social context followed by a question whose correct continuation is determined by inferring an agent's mental state, role, or the moral/normative status of an action; the corruption swaps a single word or scenario in the context (e.g., every day at the gym → at the library, won → lost, lauded → shamed) so that the correct continuation flips. Three tasks (Agent, SocRel, SocInt) are adapted from EWOK (Ivanova, Sathe, Lipkin, et al., 2025), the rest are created by us. Like for physical reasoning, those created by us were based on templates generated by Claude Opus 4.5 and manually verified by us.

**Agent (N = 945).** The agent task tests belief attribution: whether an agent should be inferred to believe or doubt a proposition given perceptual evidence. The corruption swaps a context word (e.g., inside ↔ outside) so the evidence supports the opposite belief. Original: "Ali is in the bakery. Ali sees the candle inside. Do you think that Ali doubts that the candle is in the bakery or do you think that Ali believes that the candle is in the bakery?" Alternative: "… Ali sees the candle outside. …" (correct continuation flips from Ali believes that the candle is in the bakery to Ali doubts that the candle is in the bakery)

**Desires (N = 1,080).** The desires/goals task tests goal inference from behavior. The corruption swaps the named location or activity, flipping the inferred desire. Original: "Ali goes to the gym every day. What does Ali want?" Alternative: "Ali goes to the library every day. What does Ali want?" (correct continuation flips from Ali wants to exercise to Ali wants to study)

**PrimEmo (N = 1,295).** The primary-emotions task tests inference of basic emotions (joy, sadness, anger, fear, etc.) from a short event description. The corruption swaps the valence-bearing word in the event. Original: "Ali just won the competition. How does Ali feel?" Alternative: "Ali just lost the competition. How does Ali feel?" (correct continuation flips from Ali feels joyful to Ali feels sad)

**SecEmo (N = 1,600).** The secondary-emotions task tests inference of socially complex emotions (pride, shame, guilt, envy, etc.) that depend on the actions of one agent toward another. The corruption swaps the social action word. Original: "Ali lauded Wei in front of the class. How does Wei feel?" Alternative: "Ali shamed Wei in front of the class. How does Wei feel?" (correct continuation flips from Wei feels proud to Wei feels ashamed)

**EmotionFS (N = 2,000).** The emotion-fewshot task tests emotion inference using a two-shot in-context template with two labeled exemplar contexts followed by a target context; the corruption replaces the target context with one of opposite emotional valence. Target — Original: "Mei hosted a party but nobody came over." Alternative: "Mei saw the northern lights for a night." (correct continuation flips from sadness to joy)

**SocInt (N = 204).** The social-interactions task tests inference about the nature of a social interaction: whether one agent is helping or coercing another. The corruption swaps a valence word in the context (e.g., worse ↔ better). Original: "Ali is making Wei feel worse. Do you think that Ali is helping Wei or do you think that Ali is coercing Wei?" Alternative: "Ali is making Wei feel better. …" (correct continuation flips from Ali is coercing Wei to Ali is helping Wei)

**SocRel (N = 288).** The social-relations task tests inference of social role from described behavior: whether one agent is the other's boss or subordinate. The corruption swaps the directionality of the described action. Original: "Ali gives orders to Wei. Do you think that Ali is Wei's boss or do you think that Ali is Wei's subordinate?" Alternative: "Ali gets orders from Wei. …" (correct continuation flips from Ali is Wei's boss to Ali is Wei's subordinate)

**NormAPP (N = 2,000)**. The norm-appropriateness task tests social-norm judgment using a two-shot in-context template; the corruption replaces the target scenario with one of opposite normative status. Target — Original: “Luca opened a sibling’s package that was clearly addressed to them.” Alternative: “Luca gave a friend space after they said they needed time alone.” (correct continuation flips from inappropriate to appropriate)

**NormMoral (N = 2,000).** The moral-judgment task tests moral evaluation using the same two-shot in-context template as Approp, but with judgments framed as moral vs. immoral. The corruption replaces the target scenario with one of opposite moral status. Target — Original: “Ali let a colleague take the blame for a shared mistake without speaking up.” Alternative: “Ali refused to lie to cover for a friend who had done something harmful.” (correct continuation flips from immoral to moral)

### 2.5 Prompt structure

All tasks are operationalized as next-token prediction: the model receives a prompt and we measure the log-probability assigned to the target tokens. Tasks vary in the amount of scaffolding required to elicit the target continuation. Most tasks use a brief instruction prefix, suffix, or both to constrain the output format (eg., “Reply with only the chosen statement, no other words.”). Full prompt configurations for every task, including all prefixes and suffixes, are available in our code repository.

## 3. Per-task accuracy across models

Supplementary Figure 1 shows per-task both-correct accuracy for the six large instruction-tuned models in our main analysis, across all 46 tasks and four domains. Both-correct is defined as in the main analysis: the fraction of minimal pairs on which the model assigns higher likelihood to the original continuation given the original input and higher likelihood to the alternative continuation given the alternative input. Chance-level accuracy is 0.25.

Across the 6 models and 46 tasks, accuracy is uniformly high. 45 of the 46 tasks have at least three models exceeding the 60% both-correct threshold used as the inclusion criterion in our main analysis, and 29 of 46 are passed by all six models. Per-domain mean accuracy is roughly balanced across the four domains (Language 0.84 ± 0.06, Formal 0.85 ± 0.08, Physics 0.81 ± 0.08, Social 0.83 ± 0.07). The single task that does not meet the threshold for at least three models is one Formal Reasoning task (Mul3-Vrb, three-operand multiplication and division in verbal format), which only Qwen2.5-72B passes (at 69% both-correct); all other tasks have broad cross-model coverage. Per model, task coverage ranges from 35 of 46 (OLMo-2-32B) to 46 of 46 (Qwen2.5-72B).

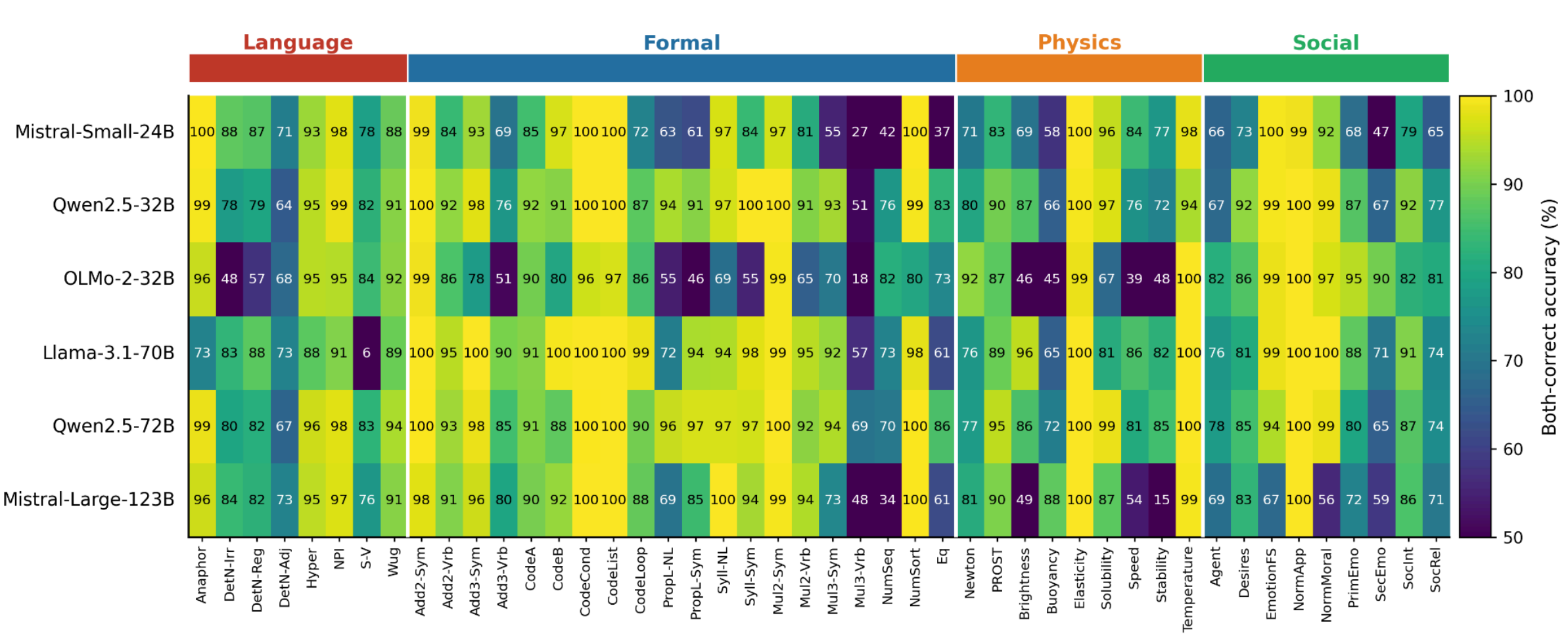

***Supplementary Figure 1.*** **Per-task both-correct accuracy across the six large instruction-tuned models.** Each cell shows the both-correct rate for one model (rows) on one task (columns), where both-correct is the fraction of minimal pairs on which the model assigns higher likelihood to the original continuation given the original input and higher likelihood to the alternative continuation given the alternative input (chance = 0.25). Tasks are grouped and color-coded by domain along the top: Language (red), Formal (blue), Physics (orange), Social (green). Cell color encodes accuracy from 50% (dark purple) to 100% (yellow); cells below 50% are clipped at the colorbar floor. Numerical values are printed in each cell. All but one of the 46 tasks reach the 60% inclusion threshold for at least three models; the one exception (mul_div_3op_verbal, in Formal) is passed only by Qwen2.5-72B.

## 4. Per-model and per-domain overlap and ablation results

### 4.1 Per-model breakdown

The main analysis reports overlap and ablation results averaged across the six large instruction-tuned models. Supplementary Table 1 gives the per-model breakdown to confirm the modular pattern is present in each individual model and does not only emerge through averaging.

Within-domain neuron overlap exceeds cross-domain overlap in every model. Within-domain overlap ranges from 11.1% (Mistral-Small-24B) to 14.2% (Qwen2.5-72B), and cross-domain overlap from 2.3% (OLMo-2-32B) to 3.5% (Mistral-Large-123B), giving ratios between

within- and across-domain overlap (henceforth W/C ratios) from 3.5× to 5.9× (mean 4.4 ± 0.9). The within–vs–cross difference is significant in every model (permutation-based $p < 0.0001$ in all six).

Causal ablation gives the same pattern. Ablating the top-0.1% within-domain neurons reduces accuracy by 18.2% to 26.4% across models, while ablating matched cross-domain neurons reduces accuracy by only 2.1% to 3.8%, yielding W/C ratios from 6.1× to 12.3× (mean 9.3 ± 2.7). Again the within–vs–cross difference is significant in every model (permutation p < 0.0001 in all six).

The pattern is therefore not driven by any single model. All six models, spanning four distinct families and 24B to 123B parameters, independently exhibit modular organization at both the correlational (overlap) and causal (ablation) level.

| Model | Params (B) | Overlap Within (%) | Overlap Cross (%) | Overlap W/C ratio | Overlap Perm $p$ | Ablation Within Δ acc | Ablation Cross Δ acc | Ablation W/C ratio | Ablation Perm $p$ |
|---|---|---|---|---|---|---|---|---|---|
| Mistral-Small-24B | 24 | 11.1 | 3.2 | 3.5 | <0.0001 | 0.235 | 0.038 | 6.1 | <0.0001 |
| Qwen2.5-32B | 32 | 12.4 | 2.6 | 4.8 | <0.0001 | 0.240 | 0.021 | 11.4 | <0.0001 |
| OLMo-2-32B | 32 | 13.9 | 2.3 | 5.9 | <0.0001 | 0.264 | 0.021 | 12.3 | <0.0001 |
| Llama-3.1-70B | 70 | 11.7 | 3.3 | 3.5 | <0.0001 | 0.182 | 0.028 | 6.4 | <0.0001 |
| Qwen2.5-72B | 72 | 14.2 | 3.5 | 4.1 | <0.0001 | 0.258 | 0.023 | 11.2 | <0.0001 |
| Mistral-Large-123B | 123 | 12.5 | 2.9 | 4.3 | <0.0001 | 0.231 | 0.027 | 8.5 | <0.0001 |
| **Mean ± SD** | – | **12.6 ± 1.2** | **3.0 ± 0.4** | **4.4 ± 0.9** | **all <1e-04** | **0.235 ± 0.029** | **0.027 ± 0.007** | **9.3 ± 2.7** | **all <1e-04** |

***Supplementary Table 1.*** **Per-model overlap and ablation results.** For each model we report the mean within-domain and cross-domain neuron overlap (top-0.1% positively attributed neurons, Jaccard index, %), their ratio (W/C), and the permutation-test p-value for the within-versus-cross difference (10,000 permutations). Ablation columns report the mean drop in both-correct accuracy when the top-0.1% within-domain or matched cross-domain neurons are replaced with the alternative value, their ratio, and the corresponding permutation p-value. The bottom row gives the cross-model mean ± standard deviation. All six models independently show significant within-domain modularity at both the correlational (overlap) and causal (ablation) level.

## 4.2 Per-domain breakdown

The modular pattern is equally clear when the analysis is broken down by individual cognitive domain (Supplementary Table 2). Within-domain neuron overlap exceeds cross-domain overlap in all four domains, with within-domain overlap ranging from 8.6% (Social) to 14.2% (Formal) and cross-domain overlap ranging from 1% (Language) to 4.7% (Physics). The within-versus-cross difference is significant in every domain (permutation test, all p ≤ 0.017; W/C ratios 2.7× to 9.9×).

Causal ablation confirms the same organization. Ablating the top-0.1% within-domain neurons produces a substantially larger accuracy drop than ablating cross-domain neurons in every domain, in both directions: when a domain's own neurons are ablated, the drop on its own tasks far exceeds the drop on other domains' tasks (D → not-D), and when other domains' neurons are ablated, a domain's own tasks are far less affected than by its own neurons (not-D → D). The within-versus-cross difference is significant in three of the four domains in both directions (Formal, Language, and Physics; all $p \leq 0.027$). For Social reasoning the effect is in the same direction in both analyses (W/C = 4.0× and 2.1×) and of comparable relative magnitude, just below the significance threshold ($p = 0.074$ and $0.146$).

Across the four domains and the three analyses (overlap, D → not-D ablation, and not-D → D ablation), the within-versus-cross asymmetry holds in every case and is statistically significant in ten of twelve, confirming that the modular organization is a property of each cognitive domain individually.

| | | Overlap | | | | Ablation | | | | | | |
|---|---|---|---|---|---|---|---|---|---|---|---|---|
| | | | | | | Within | D → not-D | | | not-D → D | | |
| Domain | $n$ | Within (%) | Cross (%) | W/C ratio | Perm $p$ | Δ acc (%) | Δ acc (%) | W/C ratio | Perm $p$ | Δ acc (%) | W/C ratio | Perm $p$ |
| Formal | 20 | 14.2 | 3.1 | 4.6 | <0.0001 | 32.0 | 3.01 | 10.6 | <0.0001 | 1.91 | 16.8 | <0.0001 |
| Language | 8 | 9.9 | 1.0 | 9.9 | 0.0048 | 20.4 | 0.46 | 44.3 | 0.006 | 0.38 | 53.9 | 0.007 |
| Physics | 9 | 12.7 | 4.7 | 2.7 | 0.0050 | 16.0 | 4.14 | 3.9 | 0.021 | 4.22 | 3.8 | 0.027 |
| Social | 9 | 8.6 | 3.0 | 2.9 | 0.0165 | 7.9 | 1.99 | 4.0 | 0.074 (n.s.) | 3.72 | 2.1 | 0.146 (n.s.) |

***Supplementary Table 2.*** **Per-domain overlap and ablation results.** For each cognitive domain, we report the mean within-domain and cross-domain neuron overlap (top-0.1% positively attributed neurons, Jaccard index, %), their ratio (W/C), and the permutation-test $p$-value for the within-versus-cross difference (10,000 permutations). Ablation columns report the mean drop in both-correct accuracy when the top-0.1% within-domain neurons are replaced with alternative values (Within Δ acc), alongside two directional cross-domain effects: ablating the domain's own neurons and measuring the drop on tasks from other domains (D → not-D), and ablating other domains' neurons and measuring the drop on the domain's own tasks (not-D → D). Within-domain overlap exceeds cross-domain overlap in every domain, and ablation effects follow the same direction throughout.

# 5. Robustness to thresholding choices

## 5.1 Modular neuron overlap is robust to thresholding choices

Throughout the main text, domain-specialized neurons are defined as the top 0.1% of positively attributed neurons for each task. To verify that our conclusions do not depend on this specific cutoff, we repeated the within-versus-cross-domain overlap analysis at four thresholds spanning two orders of magnitude: top 0.05%, 0.1%, 1.0%, and 5.0%. For each threshold we recomputed the pairwise Jaccard overlap matrix for every model, averaged the matrices across the six models, and ran the same permutation test on the within-versus-cross-domain difference (10,000 permutations of task-to-domain labels; SI Section 1).

The within-versus-cross-domain effect remains highly significant at every threshold (Supplementary Table 3, all permutation $p < 0.0001$). The 0.1% row reproduces the values reported in the main text (within-domain mean 12.9%, cross-domain mean 3.0%). The within-versus-cross ratio decreases monotonically as the threshold widens, from 5.4× at the strictest cutoff (top 0.05%) to 1.4× at the broadest (top 5.0%). This pattern is consistent with domain specificity being concentrated in the most strongly attributed neurons: as the threshold widens, the selected sets increasingly include weakly attributed, less specialized neurons that are shared across tasks and dilute the within-domain signal. Some attenuation is also mathematically expected, since Jaccard overlap between any two large sets drawn from a finite pool of neurons converges toward unity.

| Top-$X\%$ | Within (%) | Cross (%) | W/C ratio | Perm. $p$ |
|---|---|---|---|---|
| 0.05% | 12.8 | 2.4 | 5.4 | <0.0001 |
| 0.1% | 12.9 | 3.0 | 4.3 | <0.0001 |
| 1.0% | 9.8 | 4.7 | 2.1 | <0.0001 |
| 5.0% | 14.1 | 10.2 | 1.4 | <0.0001 |

***Supplementary Table 3.*** **Robustness to thresholding choice.** For each top-k% of positively attributed neurons, we report the mean within-domain and cross-domain neuron overlap (Jaccard, %), their ratio (W/C), and the permutation-test $p$-value for the within-versus-cross-domain difference (10,000 permutations on the model-averaged overlap matrix). The 0.1% row reproduces the values cited in the main text.

## 5.2 Modular ablation effects are robust to thresholding choices

The overlap analysis in Section 4.1 shows that domain-specialized neurons are shared more within domains than across domains at every threshold. In this section we test whether thresholding robustness extends to the causal contribution of these neurons. We repeated the within-versus-cross-domain ablation analysis at the same four thresholds (top 0.05%, 0.1%, 1.0%, and 5.0%) in Qwen2.5-32B-Instruct, which is the most accurate LLM in the 32B range (SI Section 3).

At every threshold, ablating within-domain neurons produced substantially larger accuracy drops than ablating cross-domain neurons (Supplementary Table 4, all $p < 0.0001$), with W/C ratios ranging from 2.6× to 12.9× across thresholds and directions. Widening the threshold naturally increases accuracy drops in all conditions, as more neurons are ablated, but the W/C ratio decreases. This pattern indicates that the neurons with the strongest domain-specific causal contributions are concentrated at the top of the attribution ranking, mirroring what we found in the overlap analysis (SI Section 5.1). As the threshold widens, the additionally included neurons carry progressively less domain-selective influence, diluting the within-versus-cross-domain asymmetry.

| | **Within** | **D** → not-D | | | not-D → **D** | | |
|---|---|---|---|---|---|---|---|
| **Top-$X$%** | **Δ acc (%)** | **Δ acc (%)** | **W/C ratio** | **Perm. $p$** | **Δ acc (%)** | **W/C ratio** | **Perm. $p$** |
| 0.05% | 15.0 | 1.16 | 12.9 | <0.0001 | 1.28 | 11.7 | <0.0001 |
| 0.1% | 18.3 | 2.04 | 9.0 | <0.0001 | 2.20 | 8.3 | <0.0001 |
| 1.0% | 29.0 | 6.20 | 4.7 | <0.0001 | 6.64 | 4.4 | <0.0001 |
| 5.0% | 43.0 | 16.64 | 2.6 | <0.0001 | 16.63 | 2.6 | <0.0001 |

***Supplementary Table 4.*** **Robustness of ablation effects to thresholding choices.** For each top-k% of positively attributed neurons in Qwen2.5-32B-Instruct, we repeated the ablation experiment described in the main text and report the mean within-domain accuracy drop (Δ acc, percentage points), the cross-domain accuracy drop, W/C ratio, and permutation-test $p$-value for both ablation directions: source task from one domain D and target task from other domains (D → not-D) and source tasks from other domains and target task from domain D (not-D → D). Permutation $p$-values are based on 10,000 permutations of task-to-domain labels. The 0.1% row reproduces the values reported in the main text.

## 6. Robustness to ablation method choices

The main analysis uses a variant of ablation in which a neuron's activation is replaced with its value on the alternative input (henceforth "counterfactual ablation", also known as *activation patching* in the literature; Hanna et al., 2026). This avoids the out-of-distribution problem that affects other intervention methods (Hanna et al., 2026). Two commonly used alternatives are zero ablation, which sets activations to zero, and mean ablation, which replaces activations with their mean across a corpus. Both can push activations out of distribution, making it difficult to distinguish a neuron's functional importance from distributional artifacts (Hanna et al., 2026). Nevertheless, to confirm that our modularity findings do not depend on the choice of intervention, we repeated the within-versus-cross ablation analysis using all three methods on Qwen2.5-32B-Instruct.

All three methods produce a consistent modular pattern (Supplementary Table 5). In every domain and both directions (D → not-D and not-D → D), ablating within-domain neurons causes a larger accuracy drop than ablating cross-domain neurons across all three ablation methods. Counterfactual ablation produces the largest accuracy drop, which is expected as it is not merely uninformative but actively misleading, driving the model toward the wrong answer.

| | | **Within** | **D** → not-D | | | not-D → **D** | | |
|---|---|---|---|---|---|---|---|---|
| **Domain** | **Method** | **Δ acc (%)** | **Δ acc (%)** | **W/C** | **Perm. *p*** | **Δ acc (%)** | **W/C** | **Perm. *p*** |
| Language | counterfactual | 19.6 | 0.4 | 49.0 | <0.0001 | 0.4 | 49.0 | <0.0001 |
| | mean | 6.2 | 0.6 | 10.3 | <0.0001 | 0.3 | 20.7 | <0.0001 |
| | zero | 8.5 | 0.7 | 12.1 | <0.0001 | 0.3 | 28.3 | <0.0001 |
| Formal | counterfactual | 32.9 | 2.7 | 12.2 | <0.0001 | 1.5 | 21.9 | <0.0001 |
| | mean | 8.0 | 0.6 | 13.3 | <0.0001 | 0.4 | 20.0 | <0.0001 |
| | zero | 6.2 | 0.7 | 8.9 | <0.0001 | 0.4 | 15.5 | <0.0001 |
| Physics | counterfactual | 14.9 | 3.6 | 4.1 | <0.0001 | 3.7 | 4.0 | <0.0001 |
| | mean | 1.8 | 0.6 | 3.0 | <0.0001 | 0.9 | 2.0 | 0.0003 |
| | zero | 3.7 | 0.6 | 6.2 | <0.0001 | 1.3 | 2.8 | <0.0001 |
| Social | counterfactual | 5.6 | 1.5 | 3.7 | 0.0009 | 3.3 | 1.7 | 0.066 (n.s.) |
| | mean | 1.9 | 0.6 | 3.2 | <0.0001 | 1.0 | 1.9 | <0.0001 |
| | zero | 1.5 | 0.6 | 2.5 | <0.0001 | 0.8 | 1.9 | 0.0010 |

***Supplementary Table 5.*** **Robustness of ablation effects to ablation method.** For each ablation method (counterfactual, mean, zero) applied to the top-0.1% positively attributed neurons in Qwen2.5-32B-Instruct, we report the mean within-domain accuracy drop (Δ acc, percentage points) alongside two directional cross-domain effects: ablating the domain's own neurons and measuring the drop on tasks from other domains (D → not-D), and ablating other domains' neurons and measuring the drop on the domain's own tasks (not-D → D). For each direction we report the cross-domain accuracy drop, the within-versus-cross ratio (W/C), and the permutation-test $p$-value (10,000 permutations of task-to-domain labels). All three methods produce a consistent modular pattern across all four domains.

## 7. Modular organization is contingent on the model's ability to solve the tasks, not a property of the data

A natural concern is whether the modular organization we report reflects a general property of the datasets and pipeline we use, or depends on the model being able to actually perform the tasks. To address this, we ran the same attribution and overlap pipeline on GPT-2-small (124M parameters), a model roughly two to three orders of magnitude smaller than those in our main analysis and one that fails most of our tasks at chance except for language tasks.

**Task inclusion.** We deliberately ran GPT-2 on the full set of 46 tasks **without** applying the 60% both-correct filter used in the main analysis. GPT-2 fails to reach this threshold on most tasks in the Physical, Social, and harder formal reasoning categories, and would be reduced to a small subset of simple Language tasks under the standard inclusion criterion. Running on the full task set lets us ask the question: when the same pipeline is applied to the same datasets on a model that cannot solve most of the tasks, does the modular structure still appear?

**GPT-2 accuracy by domain.** GPT-2 fails general reasoning tasks not at chance but well below it. We report the both-correct rate, defined as the fraction of minimal pairs on which the model assigns higher likelihood to the original continuation on the original input and higher likelihood to the alternative continuation on the alternative input, averaged across tasks within each domain. Chance for this metric is 0.25, since it requires independent binary choices on two inputs. On Language, GPT-2 reaches 0.77, roughly comparable to the 0.83 average across our six large instruction-tuned models. On the three reasoning domains, however, GPT-2 falls to 0.03 on Formal, 0.05 on Physical, and 0.09 on Social, six to ten times below chance. Inspecting GPT-2's responses, the well-below-chance performance is not due to systematically choosing the incorrect answer, but rather because GPT-2 assigns higher likelihood to the same continuation on both original and alternative inputs. It largely fails to recognize the key contrast between original and alternative questions. The large models stay well above chance in every domain (Language 0.83, Formal 0.85, Physical 0.81, Social 0.77).

**Within- vs. cross-domain modularity.** Following the same procedure as in the main analysis, we computed within-domain and cross-domain mean overlap on GPT-2 across all 46 tasks and tested the statistical significance of the difference with a permutation test. The within-domain effect was statistically detectable but small (W/C = 2.67×, $p = 0.0003$, Cohen's $d = 0.37$), well below the six large models in our main analysis, which all show $d \geq 0.68$ (all $p < 0.0001$).

**The residual effect is driven by Language.** To localize the source of GPT-2's weak modularity, we re-ran the permutation test four times, each excluding one domain. Removing Language collapsed the effect (W/C = 1.57×, $p = 0.05$, $d = 0.17$), again far below any large model in our main analysis. Removing any other domain left GPT-2's effect intact: Formal (W/C = 4.10×, $p = 0.002$, $d = 0.47$), Physical (W/C = 2.70×, $p = 0.0002$, $d = 0.38$), Social (W/C = 4.36×, $p = 0.0001$, $d = 0.44$), again, all still well below the $d \geq 0.68$ floor of the large models. The one domain that contributes any detectable within-domain organization in GPT-2 is also the one domain GPT-2 performs above chance.

Taken together, these results indicate that the modular organization we report in the main analysis is not a property of the datasets, the contrastive design, or the attribution pipeline, but is contingent on the model being able to actually perform the tasks. The same pipeline applied to the same data on GPT-2 recovers within-domain organization only in the one domain GPT-2 can perform above chance, and the effect is an order of magnitude weaker than in any large model.

## 8. Model architectures

In this section, we listed the architectural details of each of the six models we used in our experiments. All architectural specifications below are taken from each model's official Hugging Face configuration. "Neuron" refers to a unit in a transformer layer's feed-forward intermediate (MLP) sublayer; the total number of analyzed neurons per model is therefore (number of layers) x (MLP intermediate dimension).

**Mistral-Small-24B-Instruct-2501.** A ~24-billion-parameter model with 40 transformer layers, 32 attention heads per layer, and an MLP intermediate dimension of 32,768.

**Qwen2.5-32B-Instruct.** A ~32-billion-parameter model with 64 transformer layers, 40 attention heads per layer, and an MLP intermediate dimension of 27,648.

**OLMo-2-0325-32B-Instruct.** A ~32-billion-parameter model with 64 transformer layers, 40 attention heads per layer, and an MLP intermediate dimension of 27,648.

**Llama-3.1-70B-Instruct.** A ~70-billion-parameter model with 80 transformer layers, 64 attention heads per layer, and an MLP intermediate dimension of 28,672.

**Qwen2.5-72B-Instruct.** A ~72-billion-parameter model with 80 transformer layers, 64 attention heads per layer, and an MLP intermediate dimension of 29,568.

**Mistral-Large-Instruct-2407.** A ~123-billion-parameter model with 88 transformer layers, 96 attention heads per layer, and an MLP intermediate dimension of 28,672.

## 9. A qualitative inspection of modular ablation effects

We give a qualitative example to illustrate how the modular ablation effect manifests in free-form generation. We prompted Qwen2.5-32B-Instruct with a natural-language physical reasoning question (the heat-aging of a rubber band) and inspected its open-ended output under three conditions: no ablation, ablation of language neurons, and ablation of physical reasoning neurons (Supplementary Figure 2). For each domain the ablated set was the top 5% most positively attributed neurons averaged across that domain's tasks, set to zero during generation; we use a wider threshold than the main analysis (top 5% vs. top 0.1%) so that the behavioral effects are visible in a single generation. The two ablations produced opposite error profiles on the same question: ablating language neurons left the physical reasoning correct (a weaker restoring force, less complete return to length, less energy dissipated) but introduced syntactic and morphological errors, whereas ablating physical reasoning neurons preserved grammatical well-formedness but led to incorrect physical conclusions.

Question: Two identical rubber bands - one left in a hot car for several days, one kept at room temperature. When stretched and released, how do they differ in restoring force, return to length, and energy dissipated?

No Ablation

" ... the elastic restoring force of the heat -aged rubber band will be weaker compared to the non-heat-aged band. When stretched, it will not exert as strong a force ...."

" ... the heat-aged rubber band will not return fully to its original length after being stretched"

Physical Reasoning Grammar

Ablate Language Neurons

" ... within the polymer chain network that form its structure."

"...the entangment and covalent bonds within the polymer network."

"... [heat-aged] the elastic restoring force will be weaker and less effective in returning the band to its original shape"

Physical Reasoning Grammar

Ablate Physics Neurons

" Heat aging can cause cross-linking between polymer chains ... This results in a higher elastic restoring force when band is stretched."

... the heat-aged band will likely return more closely to its original length ..."

... Heat aging leads to a reduction in the amount of energy that is dissipated as heat..."

Physical Reasoning Grammar

***Supplementary Figure 2.*** **Qualitative example of the double dissociation.** Free-form output of Qwen2.5-32B-Instruct on a physical reasoning question (rubber-band heat-aging) under three conditions. With no ablation (left), the output is both physically correct and linguistically well formed. Ablating language neurons (middle) leaves the physical reasoning correct but introduces syntactic and morphological errors (red); ablating physical reasoning neurons (right) preserves grammatical well-formedness but yields incorrect physical

conclusions (red). For each domain the ablated set is the top 5% positively attributed neurons (averaged across the domain's tasks), set to zero during generation.

## 10. Modular organization is not driven by semantic similarity of the prompts

An additional concern is whether the modular neuron overlap reported in the main analysis reflects functional specialization of the neurons engaged in the relevant reasoning or linguistic computation or instead reflects semantic similarity between prompts. Tasks within a domain naturally have similar templates and partially share their vocabulary, so one question is whether any similarity measure over the prompts (e.g., word or sentence embeddings) could recover the same four-cluster structure as the neuron overlap matrix. To rule this out, we replaced the neuron overlap matrix with task-level semantic similarity matrices computed directly from the input prompts and asked whether the clustering emerges from any of them. Specifically we (i) applied the main analysis's clustering pipeline using different similarity measures and compared the Adjusted Rand Index (ARI) against the four cognitive domains, and (ii) ran a partial regression on the 1,035 off-diagonal pairs to test whether neuron overlap carries information about the four cognitive domains beyond what the semantic baseline already explains.

We considered four semantic similarity baselines spanning from shallow to rich semantic representations. TF-IDF (Salton & Buckley, 1988) is a sparse bag-of-words representation in which each prompt is described by the term frequencies of its words, weighted by inverse document frequency; TF-IDF captures vocabulary overlap but is order-insensitive and ignores anything about which words are semantically related. GloVe (Pennington, Socher, & Manning, 2014) is a static word-embedding model that places each word in a dense vector space that encodes word meanings based on the words' distributional co-occurrence patterns; we used the 100-dimensional vectors trained on Wikipedia and Gigaword. SBERT (Reimers & Gurevych, 2019) is a sentence-level transformer encoder fine-tuned with a siamese objective to produce sentence embeddings that match human semantic-similarity judgments. The fourth baseline, the semantic control that is closest to our main analysis, is the Qwen2.5-32B input-token embeddings: the static lookup table that maps tokens to vectors at the model's first (pre-attention) layer. If the four cognitive domains were already separable in the model's input-embedding space, that alone could explain the neuron-overlap result.

For all four baselines we used the original (unmodified) clean prompts from the minimal pairs, the same prompts on which neuron attribution is computed in the main analysis. For TF-IDF, each task was represented by the TF-IDF vector of its concatenated prompts. For GloVe, each prompt was represented by the mean of its word vectors, and each task by the mean of its prompt vectors. For SBERT, each prompt was encoded into a 384-dimensional sentence embedding by the pretrained model and each task was represented by the mean of its prompt embeddings. For the Qwen input-token embeddings, each prompt was represented by the mean of the input-embedding vectors of its tokens, and each task by the mean of its prompt vectors. Pairwise task similarity was then measured by cosine similarity, yielding a $46 \times 46$ similarity matrix for each baseline. We then (i) applied the main analysis's clustering pipeline to each similarity matrix and compared the resulting four-cluster solution against the four cognitive domains using the Adjusted Rand Index (ARI), and (ii) ran a partial regression on the 1,035 off-diagonal task pairs to test whether neuron overlap carries domain information beyond what the lexical baseline already explains.

The neuron overlap matrix obtains ARI = 0.78 (permutation $p < 0.001$), substantially above every semantic baseline: SBERT 0.39 ($p < 0.001$), GloVe 0.36 ($p < 0.001$), TF-IDF 0.12 ($p = 0.01$), and Qwen2.5-32B input-token embeddings 0.04 ($p = 0.17$, *n.s.*). Thus, none of the (lexical) semantic baselines, including the input token embeddings from one LLM used in the main analysis, recover the four-domain structure as well as the neuron overlap matrix obtained with attribution patching.

In a partial regression analysis, we modeled the standardized pairwise neuron overlap value as a linear function of a same domain indicator (1 if both tasks belong to the same cognitive domain, 0 otherwise) and the corresponding standardized pairwise semantic similarity from one of the four baselines. We tested the same-domain coefficient with a 10,000-iteration label-permutation test in which the assignment of the tasks to the relevant domains was randomly shuffled and the regression refitted. The coefficient associated with the domain remained positive and significant in every case: $\beta = 0.38$ (TF-IDF); $\beta = 0.47$ (GloVe); $\beta = 0.14$ (SBERT); $\beta = 0.42$ (Qwen input-token embeddings; all $p < 0.001$). The domain effect on neuron overlap therefore persists after controlling for every semantic baseline tested, including pretrained sentence embeddings and the main-analysis model's own input-token embeddings, confirming that the modularity effects reported in the main text reflect domain-specific computations above and beyond semantic similarity.